\input epsf

\newcount\AbbCount
\newcount\hilf
\long\def\AbbildEPS#1#2#3%
 {\begin{figure}[htbp]
  \vspace{5mm}
  \epsfysize=#3 
  \hfil\epsfbox{#1.eps}\hfil
  \caption{#2}
  \label{#1}
  \vspace{2mm}
  \end{figure}}%

\documentstyle[12pt,a4,epsf]{article}

\date{}

\def\kw#1{{\it #1}}
\def\fett#1{{\bf #1}}
\font\nineit=cmti9

\begin{document}
\pagestyle{empty}

\title{\Large\bf Domain and Language Independent \\
Feature Extraction for 
Statistical Text Categorization}

\author{Thomas Bayer, Ingrid Renz, Michael Stein, Ulrich Kressel\\
	Daimler Benz AG - Research and Technology\\
	Wilhelm-Runge-Str.~11, 89081 Ulm, Germany\\
	e-mail:\{bayer, renz, kressel\}@dbag.ulm.daimlerbenz.com\\
	        }

      \maketitle

\thispagestyle{empty}

\parindent0cm
{\Large \bf Abstract}
\medskip
\renewcommand{\baselinestretch}{0.8}
\small\normalsize                 

{\small A generic system for text categorization is presented which uses a 
representative text corpus to adapt the processing steps: feature extraction, 
dimension reduction, and classification. Feature extraction automatically 
learns features from the corpus by reducing actual word forms using statistical 
information of the corpus and general linguistic knowledge. The dimension of 
feature vector is then reduced by linear transformation keeping the essential 
information. The classification principle is a minimum least square approach 
based on polynomials. The described system can be readily adapted to new 
domains or new languages. In application, the system is reliable, fast, and 
processes completely automatically. It is shown that the text categorizer works 
successfully both on text generated by document image analysis - DIA and on 
ground truth data.}

\renewcommand{\baselinestretch}{1}
\small\normalsize                 

\medskip
\medskip
\parindent0cm
{\Large \bf 1 Introduction}
\medskip

Text categorization is an important task in handling electronic text 
automatically and assigns a pre-defined category (message type) to a text 
consisting of a sequence of words. Possible applications of text categorization 
systems are information filtering and information retrieval. Furthermore, text 
categorization is necessary to reduce the complexity for subsequent 
natural-language processing.

\parindent0.5cm
The text categorizer presented is embedded in a system which handles text 
documents. The goal of the system (see \cite{bayerHB}) is to extract the 
information from paper documents in order to support sub-sequent processing. 
The analysis starts with document image analysis - DIA (see \cite{bayerDAS}) 
and returns an electronic text which contains a certain amount of errors, due 
to segmentation and character recognition errors. Based on this electronic 
text, the document is assigned to a certain message type by the text 
categorizer. 

In the following section, the system for categorization is regarded as a task 
of statistical pattern classification based on a training corpus. The 
subsequent two sections describe the features extracted from the text, the 
feature transformation, and the classification principles applied. Finally, 
categorization results are discussed for an exemplary task.

\medskip
\medskip
\parindent0cm
{\Large \bf 2 System Overview}
\medskip

In the approach presented here, text categorization is regarded as a task of 
statistical pattern classification task (see \cite{schueHB}) where each text 
represents one pattern and a text category represents a class. Hence, a {\it 
training phase} and an {\it application phase} must be distinguished. During 
training text samples are observed which define the feature set (text 
descriptors), the rules for dimension reduction and classification; during 
application the text object is mapped to its class using these sources.

\parindent0.5cm
Since feature extraction and classification are adapted by statistical 
observations (training corpus), this architecture leads to a {\bf generic 
categorization system} which is not designed to solve a specific task but 
consists of tools which are trained for each new task arising. Hence, the 
categorization system is {\bf domain and language independent}.

A consequence is that an adapted categorizer is fault tolerant to high degree 
if the errors which occur in the text are systematic such as DIA errors. Thus, 
corrupted DIA output is categorized equally well as error-free text. Another 
consequence is that the adaptation of a new categorizer costs only computation 
time. Hardly any manual effort must be spent. The only prerequisite the system 
has is that a representative set of training texts (text corpus) along with 
their class membership (labels) is available.

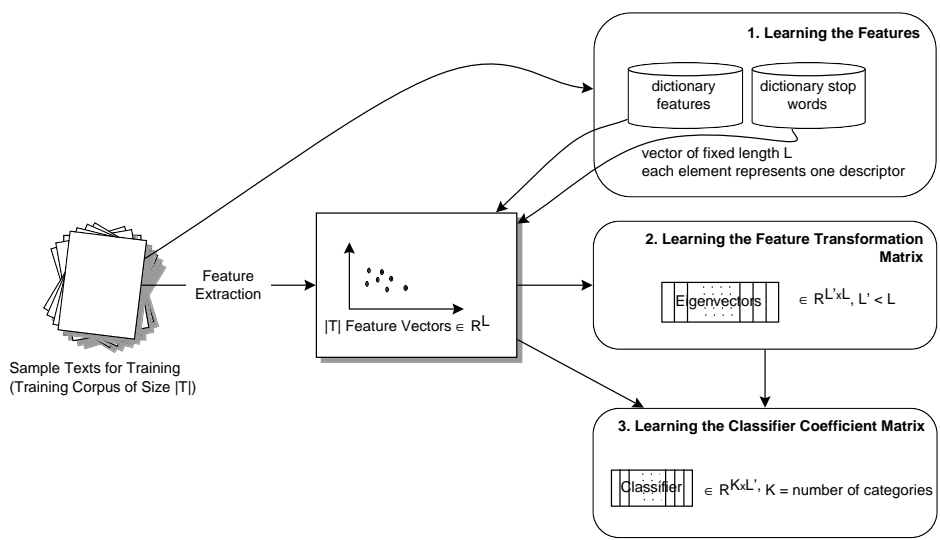
\AbbildEPS{trainingOverview}{All components of the generic system are 
adapted: dictionaries collect features which generate the eigenvector matrix; 
both create the classifier matrix.}{70mm} 

The three steps of the system: {\it Feature Extraction, Feature 
Transformation}, and {\it Classification} are trained in three steps as 
sketched in Fig.~\ref{trainingOverview}. First, relevant features 
(descriptors) are acquired from the training corpus (see sect.~3), 
automatically generating a list of stop words and a dictionary of features 
represented as a vector of fixed length $L$. Using these lists, each text of 
the training corpus is converted to its feature vector. Second, these features 
are used to create the transformation matrix used to map the feature vectors 
from dimension $L$ to a lower dimensional vector space $L'$ and third, the same 
features and the matrix are used to adapt the coefficients of the classifier 
(both see sect.~4).

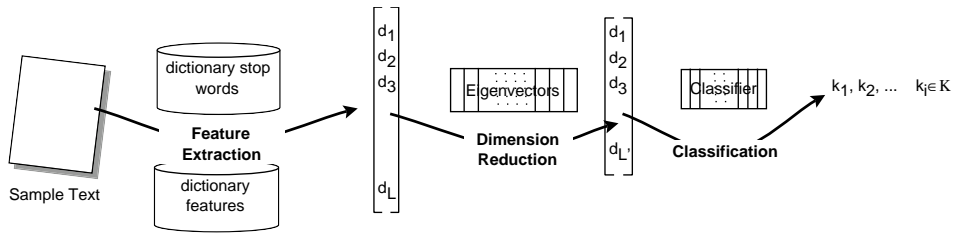
\AbbildEPS{applicOverview}{Text categorizer: all learned sources are 
used to map a text to a category.}{30mm} 

During application, a text is classified to its category as sketched in 
Fig.~\ref{applicOverview}. Stop words and descriptors are used to 
generate the feature vector of dimension L, which is transformed and classified 
by using the two learned matrices. The classification results in vector of 
length $K$, where each component $k_i$ denotes the a-posteriori probability of 
vector $v$: $k_i = p(k_i|v)$. Currently, the decision rule, which is also used 
in the sect.~5, is forced recognition, meaning that class $i$ with the maximum 
$k_i$ is selected.

\medskip
\medskip
\parindent0cm
{\Large \bf 3 Feature Extraction}
\medskip

Feature extraction defines which linguistic parts (character strings, 
morphemes, word forms, phrases) are useful features for the classifying task 
and how these features are gained from the texts to be categorized. To be 
useful for the classifying task means that features are typical for the 
category which is assigned to the text and not specific for the text itself.

\parindent0.5cm
In previous works, several approaches and aspects of feature extraction have 
been described. \cite{lewis} discusses which parts of text are suited as 
features and evaluates words and phrases with different frequencies in 
different kinds of English text. Other work, mostly for texts of morphological 
richer languages than English, considers smaller parts of text like morphemes 
(gained by linguistic lexicon-based analysis see \cite{hoch}) or n-grams 
(gained by statistical computations see \cite{cavnar}). N-grams have the 
advantage that they can be easily computed, but it is difficult to select and 
weight them for classification tasks. This is even more problematic if DIA 
errors drastically increase their number. On the other hand, linguistically 
analyzed morphemes are well defined but a respective lexicon must be 
accommodated. A further aspect of feature extraction is whether domain-specific 
features are superior to general ones (discussed and supported by \cite{apte}).

According to \cite{apte}, we propose the use of features which are specific for 
the actual categorization task. A generic procedure acquires these features 
using the training corpus and builds dictionaries of features and of stop 
words. This approach is similar to \cite{riloff} (generation of a domain 
specific lexicon using a corpus of training texts) and \cite{black} (reduction 
of complex morphemes using simpler morphemes detected in the training corpus), 
since a corpus of training texts is the base for the acquisition of knowledge 
which is necessary for a specific task. The main advantage of this approach is 
that characteristics of the actual categorization task like a specific 
vocabulary or DIA errors are automatically integrated into the categorizer.

Therefore, the basic idea of our feature extraction is that generic procedures 
reduce rather autonomously actual word forms of the corpus to features and 
result in gathering these features into a dictionary. The reduction is able to 
consider statistical information inherent in the corpus and restrictions of 
simple but appropriate general linguistic knowledge. Further knowledge bases 
are not necessary.

In the following, the steps needed to build and to use task-specific 
dictionaries are described in detail. For illustrative purpose, we refer to the 
exemplary categorization task of DIA'ed abstracts of technical reports in 
German. This example handles a language with complex morphology and word forms 
with recognition and segmentation errors. The computational complexity of this 
learning depends on the number of different word forms ($WF$) in the corpus and 
is maximally $O(WF^2)$.

\medskip
\medskip
\parindent0cm
{\large \bf 3.1 Corpus-Based Learning of Dictionaries: from Words to Features}
\medskip

In order to learn task-specific dictionaries of features and of stop words, all 
word forms of the corpus together with their frequencies are collected into a 
list. Here, a word form is defined as a character string between blanks without 
punctuation marks. It includes also forms with recognition and segmentation 
errors. In the following, the steps necessary to transforms this list of word 
forms into a dictionary of features are described:

\medskip
\parindent0.5cm

{\bf Statistical determination of stop words}

\AbbildEPS{stopwords}{Exemplary stop words of the task {\it abstracts 
of technical reports in German}}{50mm}

Stop words are defined according to their frequency in the corpus and a given 
threshold which has to be set and inspected. Fig.~\ref{stopwords} lists 
some stop words of our categorization task. Stop words include the typical 
function words of German (articles, prepositions, auxiliaries like {\it der, 
die, auf, bei}, etc.). Since all texts belong to a domain with specific 
vocabulary, stop words also include domain-specific terms which are equally 
distributed in all categories (typical words of abstracts which are independent 
of the specific category like {\it arbeit - 'work', bericht - 'report', 
beschreiben - 'describe'}). Even words containing frequent DIA errors are 
considered as stop words (e.g.\ the wrongly recognized {\it dio} instead of 
{\it die}). All stop words are collected in a dictionary of stop words and 
eliminated from the list of word forms.

\medskip

{\bf Setting of linguistic parameters}

A small number of linguistically motivated parameters are needed to model 
language-specific characteristics which a categorizer has to respect. First, 
they define the character set of the texts, distinguished into vowels and 
consonants. Then, they represent orthographic conventions (e.g.\ German 
character strings like {\it sch} or {\it ck} express only one consonant and are 
therefore treated as one character). Both definitions are necessary to restrict 
the minimal form of a feature: it consists of 3 or more characters and at least 
one of the characters has to be a vowel. This minimal form restricts the 
further reduction of word forms to features because the remaining parts of 
every split or shortening has to agree with this definition --- otherwise, the 
iterative procedure of splitting word forms could terminate with the alphabet. 

\medskip

{\bf Statistical determination of prefixes and suffixes}

\AbbildEPS{affixes}{Prefixes and suffixes of the task {\it abstracts of 
technical reports in German}}{50mm}

Before splitting the word forms, typical affixes of the training corpus are 
statistically computed according to their frequency and a given threshold which 
has to be manually set and inspected. Fig.~\ref{affixes} shows the 
results for our categorization task. Domain-specific content words which are 
frequent parts in composite words and which are equally distributed in all 
categories like {\it verfahren - 'procedure'} are treated as suffixes.

\medskip

{\bf Iterative splitting of complex word forms using simpler forms}

\AbbildEPS{splitting}{Exemplary splitting of the task {\it abstracts of 
technical reports in German}}{200mm}

This step is the most expensive and transforms the list of word forms into the 
list of possible features. By iterative pattern matching, complex word forms 
are split into smaller ones. 
Fig.~\ref{splitting} illustrates how the list of word forms is 
transformed: in the first cycle {\it halbleitertechnik - 'technology of 
semi-conductors', 'solid state technology'} is split into {\it halbleiter} and 
{\it technik} (and removed from the list) since {\it halbleiter} is part of the 
list and both {\it halbleiter} and {\it technik} are conform to the linguistic 
parameters. The frequency of {\it halbleitertechnik} is added to the frequency 
of {\it halbleiter}) and is set as the frequency of the new list item {\it 
technik}. The next cycle splits {\it halbleiter} since {\it halb} is part of 
the list and results in the new list item {\it leiter}. If no further split is 
possible, the procedure terminates.

This procedure exploits the morphological regularity that parts of composite 
words exist as simple forms. If both, the complex form and a simpler one which 
is part of it, are members of the list of word forms, the complex form is 
divided into the simple one and the remaining character string (which has to 
respect the linguistic parameters). According to language specific properties, 
several cycles of splitting complex character strings into simple ones are 
necessary for German word forms whereas only a few are needed for English ones.

\medskip

{\bf Subsequent elimination of suffixes and prefixes}

In order to eliminate character strings which have mostly formative and no 
content function, the computed suffixes and prefixes are used to shorten the 
forms of the list as long as the linguistic parameters are respected. Since 
this shortening has the effect that different forms become equal (augmenting 
their frequency), the list contains less forms.  

A German morphological characteristic is that formative elements exist between 
the parts of a composite word. The most prominent example is the {\it 
Fugenelement 's'}, e.g.\ in {\it anfangswert - 'start value'}. A further rule 
matches two forms of the list, if they are the same except that one of them 
starts or ends with such a formative element.

Fig.~\ref{finalforms} shows the final (split and shortened) forms 
together with their resulting frequencies.

\AbbildEPS{finalforms}{Exemplary features of the task {\it abstracts of 
technical reports in German}}{60mm}

A final form of the list can be interpreted as lying between n-grams (3 or more 
characters) and word stems according to text quality and language. If the DIA 
results in oversegmented character strings, these faulty-segmented strings 
further split word forms and the features are similar to n-grams. If the text 
quality is good and the text contains only a few DIA errors, the features in 
morphological rich languages as German correspond to word stems.

\medskip

{\bf Selecting features according to a given frequency threshold}

Finally, the genuine features have to be selected from the list of final forms 
according to their frequency.  Generally, all forms that occur only once or 
twice are irrelevant for the following classification, therefore, the threshold 
has to be higher than 2. For actual categorization tasks, different thresholds 
have been set between 3 and 19. 

The selected features are then stored in a dictionary of features which 
together with the dictionary of stop words constitutes the knowledge base of 
the following step.

\medskip
\medskip
\parindent0cm
{\large \bf 3.2 Application of Dictionaries: from Texts to Vectors}
\medskip

Using the acquired task-specific dictionaries, the texts are transformed into 
feature texts. First, the domain specific stop words are eliminated according 
to the stop word dictionary. Then, the remaining word forms are replaced by 
features of the feature dictionary if these features are part of the word 
forms, otherwise the word forms are deleted. This transformation of a text 
(DIA'ed abstract of a technical report in German) is shown in 
fig.~\ref{ex_text} .

\AbbildEPS{ex_text}{Example of an DIA'ed text together with its 
corresponding feature text}{90mm}

\parindent0.5cm
Generally, both dictionaries are rather small (the number of stop words lies 
between 20 and 200, the number of features between 1000 and 10000) and 
therefore, the completely automatic matching procedure is very fast.

Since the number of all features given by the dictionary is a-priori known, 
they are  represented by a feature vector of fixed length $L$. In the 
experiments, binary feature vectors have been used. Besides binary values, 
frequency scores can be computed, such as inverse document frequency etc. Tests 
have shown that the recognition accuracy is not much affected.

\medskip
\medskip

Summarizing the main properties of our feature extraction:

\begin{itemize}
\item Task-specific dictionaries of features and stop words are acquired from 
corpus and applied to texts. This approach allows an easy adaptation to new 
domains and languages.
\item Features can be interpreted as lying between n-grams and word stems. 
Domain-specific content words, language-specific function words, and affixes 
which have only syntactic or domain overlapping meaning are ignored.
\item The resulting categorizer is fault tolerant since the features are 
adapted to DIA input.
\item The generic procedure operates only on the corpus of training texts. 
There is no need for expensive (lexical) resources or further knowledge bases.
\end{itemize}

\medskip
\parindent0cm
{\Large \bf 4 Classification}
\medskip

Classification is here considered from the statistical point of view. Given a 
training set of objects $o_i, i = \{1,\ldots N\}$ along with their class label 
$k \in \{1,\ldots,K\}$, a classifier is constructed. The feature vector $v_i 
\in R^L$ to each object $o_i$ is calculated as described in the previous 
section with a dimension ranging from 1000 to 10000 in our applications. Before 
adapting the classifier by the set of $v_i$, the dimension L is reduced to a 
reasonable small number $L'$ of several hundreds for two reasons: First, there 
is a strong relationship between the dimension of the feature space $L$ and the 
required number $N$ of training samples; the higher $L$ is, the more training 
examples must be provided in order to avoid overfitting to the training set. 
Second, such a high dimension would cause high computing effort both for the 
adaptation phase and for the classification. 

\parindent0.5cm
Hence, before constructing the classifier, the dimension of the vector space is 
reduced by using the same training set of objects. The resulting pairs $(v'_i, 
k_i), v'_i \in R^{L'}$ are then the basis for constructing the classifier. Both 
processes are described in the following.
\medskip

\fett{Dimension Reduction} One well known method to reduce the vector space 
$R^L$ is the principal component analysis (PCA) which is based on the 
eigenvalues and eigenvectors of the covariance matrix $C = 1/N\sum_{i = 
1}^{N}(v_i - \mu)(v_i - \mu)^T, \mu = 1/N\sum_{i = 1}^{N}v_i$. \footnote{we 
used our own software implementations for eigenvector analysis and linear 
regression.} The $L$ eigenvectors constitute a orthonormal basis B each vector 
$v_i$ can be represented in: $v_i' = B^Tv_i$. The essential property of this 
linear transformation is the following: the PCA minimizes the euclidian 
distance between $v_i$ and $v_i'$ (also called the {\it reconstruction error}) 
if $v_i'$ is a linear combination of the $L'$ eigenvectors belonging to the 
$L'$ greatest eigenvalues, $L' < L$.

In the experiments described below, $L$ has been in the range of $2500$ and 
$L'$ has been selected from the set $50, 100, 200, 500$. 
Fig.~\ref{patError} displays the loss of information, i.e.~the 
reconstruction error for different $L'$ (numbers of eigenvalues selected), and 
motivates the selection of these 4 values. For example, using the first 50 
eigenvectors ($L' = 50$), approx.~70\% of the information is lost; it does not 
seem reasonable to reduce the number further since the transformed vectors 
$v_i'$ tend to become meaningless. Using the first 500 eigenvectors, 90\% of 
the information is preserved. Selecting more than 500 coefficients does not 
seem to yield additional benefits since most of the information is already 
available.

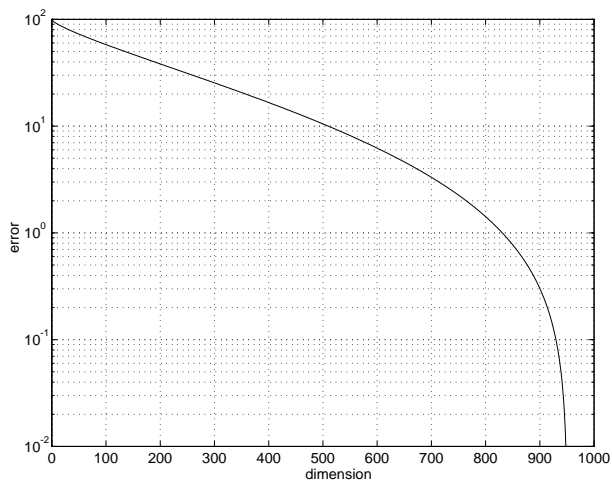
\AbbildEPS{patError}{Reconstruction error in percent for different 
number of eigenvectors.}{70mm}

An alternative approach,the SVD (singular value decomposition) is also applied 
to reduce the dimensionality of feature vectors. The SVD is not based on the 
covariance matrix $C$ but on the matrix $A$ of dimension $N\times L$ 
represented by the feature vectors $v_i$. It also minimizes the reconstruction 
error between $v_i$ and $v_i'$ if only a subset of the orthonormal vectors of 
the decomposed matrix are used. Hence, PCA and SVD are closely related, but not 
identical. Usually, such techniques are used for \kw{Latent semantic indexing} 
(see \cite{deerwester}) since each component of the $v_i'$ in the system of the 
eigenvectors can be interpreted as a (linear) combination of vector elements in 
the original feature space.

Note that the principal component analysis is class independent since each 
$v_i$ regardless of its class is transformed by the same matrix $B$. A 
different linear transformation for the same purpose of reducing L 
significantly is the linear discriminant analysis which is class specific and 
has been used by \cite{karlgren}.
\medskip

\fett{Classification} The final step in text categorization is the mapping of 
an $L'$-dimensional feature vector (measurement space) into one of $K$ classes 
(decision space). The classification principle employed here is functional 
approximation based on polynomials. The $L'$ elements of $v_i \in R^{L'}$ are 
combined by a polynomial function $x: v \rightarrow x(v)$,  resulting in 
multiplicative combination of the elements. For example, a second order 
polynomial function $x$ generates the $L'^2$ quadratic and $L'$ linear 
polynomials of each element $v^j_i$ of $v_i$. Mathematically, the polynomial 
classifier is defined as $d(v) = A^T \times x(v)$, where $A \in R^{K\times X}$ 
is the coefficient matrix to be adapted and $X$ the dimension of the range of 
the function $x$. The coefficients are calculated by minimizing the mean-square 
error between the estimation $d(v)$ and the true value $y$ describing the class 
membership of $v$:

$$E\{ |A \times x(v)-y|^2 \} = Minimum. $$

$E\{\ldots\}$ denotes the mathematical expectation and $y$ the desired target 
vector which is a unity vector having the '1' at the k-th position if $v$ 
belongs to class k. In the optimization problem above $A$ is computed by linear 
regression\footnotemark[1] assessing a training sample of size $N$ of pairs 
$(v_i, y_i)$. It can be shown that the k-th element of $d(v)$ estimates the 
a-posteriori probability $p(k|v)$. For a detailed description of the polynomial 
classifier design see \cite{schueHB}. 

Depending on the dimension $L'$ and on the training set size $N$, a linear or 
higher order classifier can be constructed. The construction of a higher order 
polynomial classifier -- which in general gains a higher recognition accuracy 
than a linear one  -- is only reasonable when the dimension $L'$ is small with 
respect to $N$ since the number of parameters to be adapted by the training 
samples grows with ${L' \choose p}$, $p$ being the order of the polynomial. 
Hence, a higher order (second) polynomial classifier is only appropriate if $L' 
< 100$ and the number of samples $N_k > 1000$ of each class $k$. 

In the current applications, linear classifiers have been adapted for the 
different sizes of $L'$. The current linear classifier is identical to the LLSF 
(linear least square fit) classifier described by Yang \cite{yang92} and 
\cite{yang95}. However, the mathematical principle is different in general if 
higher order polynomials are used. In this case, a non-linear function 
(e.g.~quadratic polynomial) maps the feature space to the decision space 
yielding better separation of classes in the decision space.

\medskip
\medskip
\parindent0cm
{\Large \bf 5 Results}
\medskip

One exemplary domain in which the text categorizer has been applied are 
abstracts of technical reports. Every technical report (total number: 1144) 
belongs to one of six classes: solid state physics, telecommunications, 
material science, information processing, opto-electronics, and pattern 
recognition. The cardinalities of the classes are approximately equal. It is a 
rather hard categorization task since some classes are closely related, 
e.g.~information processing and pattern recognition, and some texts contain 
mixed subjects - even persons who labelled the abstracts had difficulties. The 
reports were transformed by DIA into ASCII resulting in a word accuracy of 
83.6\% (details about the algorithms can be found in \cite{bayerDAS}).Then, all 
texts were manually corrected resulting in a second input set, the ground truth 
data. With our experiments, we have examined the following variations:
\begin{description}
\item [Feature extraction] In order to evaluate the approach presented here 
(learned features), we also extracted feature sets by the method of tri-grams 
(see \cite{cavnar}) and for the corrected texts by morphological analysis with 
a complete lexicon (see \cite{hoch}). All feature sets have approximately the 
same size of 2500 features.
\item [Feature transformation] Since the vector length also influences the 
categorization result, the principal component analysis results in vector 
lengths of 50, 100, 200, and 500. 
\end{description}
\parindent0.5cm
In the following two tables, the error rates of the categorizer under the 
condition of forced recognition (i.e.~the categorizer always assigns one 
category to one text without the possibility to reject texts or to assign 
several categories) are shown. The first table contains results for texts 
generated by DIA, whereas the last table shows the results of the ground truth 
data. 

The first categorizer in table \ref{c1} is based on 950 training texts and 140 
test texts. The lowest error rate is gained with a dimension space of 500 
features. The comparison of the tri-grams and the learned features shows that 
the tri-gram approach needs only 200 dimensions in order to have its best 
result whereas our approach needs 500 dimensions and is better than the best 
categorization result of the tri-grams. 

\begin{table}[htbp]
\begin{center}
\begin{tabular}{||r||r|r||} \hline\hline
vector length & tri-gram features & learned features \\ \hline\hline
 50 &  28.8{\%} & 25.2{\%} \\
100 &  27.3{\%} & 20.9{\%} \\
200 &  22.3{\%} & 20.1{\%} \\
500 &  26.6{\%} & 17.3{\%} \\ \hline\hline
\end{tabular}
\end{center}
\caption[tb-comparison tri-learn]
        {Results on test texts for two different feature sets resulting from 
DIA input}
\label{c1}
\end{table}

The second categorizer in table \ref{c2} is adapted in order to compare our 
learned features with morphological features. In order to apply the morphology 
system, the texts were transformed into ground truth data and a complete 
dictionary for these texts was developed. Here, the number of training texts is 
1004, the number of test texts 140. Again, the 500-dimensional vector space in 
combination with the learned features has the lowest error rate. An explanation 
for the surprisingly high error rate of morphemes could be that a statistical 
classifier is not appropriate for this kind of features.

\begin{table}[htbp]
\begin{center}
\begin{tabular}{||r||r|r|r||} \hline\hline
vector length & morphological features & tri-gram features & learned features 
\\ \hline\hline
 50 & 48.3{\%} & 27.9{\%} & 30.7{\%} \\
100 & 44.6{\%} & 26.4{\%} & 28.6{\%} \\
200 & 41.5{\%} & 23.6{\%} & 23.6{\%} \\
500 & 38.2{\%} & 22.9{\%} & 21.4{\%} \\ \hline\hline
\end{tabular}
\end{center}
\caption[tb-comparison morph-tri-learn]
        {Results on test texts for different feature sets resulting from ground 
truth data}
\label{c2}
\end{table}

It has to be pointed out that in every categorization task, the learned 
features extracted by our statistical approach result in the best recognition 
rates. Interestingly, the best error rate on the DIA input is slightly better 
than the best on the ground truth data. This example indicates that errors does 
not deteriorate the recognition performance.

Finally, the major property of the text categorizer presented here should be 
stressed again, which is the minimal manual effort to adapt the complete system 
to new categorization tasks.

\medskip
\medskip
\parindent0cm
{\Large \bf 6 Future Work}
\medskip

Currently, a drawback of the classifier is that text objects must correspond to 
exactly one text category; mathematically, the target vector is a unit vector 
with the '1' at the class index. However, often a text can be assigned to more 
than one class. In the near future, the range of the target vectors is extended 
to the range of real values, more precisely between $[0;1]$. Each non-zero 
value denotes then to what degree a text can be assigned to this class. The 
mapping can then be approximated more precisely, yielding higher recognition 
scores. 

\parindent0.5cm
A second future topic is to reduce the manual effort to a minimum. Currently, 
several parameters during generation of the dictionaries are set by inspection 
of intermediate results, e.g.~the thresholds for stop words and for the 
decision what descriptors are selected as features. These thresholds shall be 
replaced by statistical observations.

\renewcommand{\baselinestretch}{0.8}
\small                


\begin{thebibliography}{99}
\setlength{\parskip}{0pt}

\bibitem{apte} C.~Apt\'e, F.~Damerau, S.~Weiss: Towards Language Independent 
Automated Learning of Text Categorization Models, {\nineit Proceedings of 
SIGIR}, 1994
\bibitem{bayerHB} T.A.~Bayer, U.~Bohnacker, I.~Renz: Information Extraction 
From Paper Documents, to appear in: P.S.P.~Wang, H.~Bunke (eds.), Handbook of 
Optical Character Recognition and Document Image Analysis, 1996
\bibitem{bayerDAS} T.A.~Bayer, U.~Bohnacker, H.~Mogg-Schneider: InfoPortLab -- 
An Experimental Document Analysis System, {\nineit Proceedings of the 1st 
Workshop on Document Analysis Systems}, Kaiserslautern, 1994
\bibitem{black} A.~Black, J.~Plassche, B.~Williams: Analysis of Unknown Words 
through Morphological Decomposition, {\nineit Proceedings of European Chapter 
of the Association for Computational Linguistics}, 1991
\bibitem{cavnar} W.B.~Cavnar, J.M.~Trenckle: N-Gram-Based Text Categorization, 
{\nineit Proceedings of the Symposium on Document Analysis and Information 
Retrieval}, Las Vegas, 1994
\bibitem{deerwester} S.~Deerwester, S.T.~Dumais, G.W.~Furnas, T.K.~Landauer, 
R.~Harshman: Indexing by Latent Semantic Analysis, {\nineit Journal of the 
American Society for Information Science}, 41(6), 1990
\bibitem{hoch} R.~Hoch: Using IR Techniques for Text Classification in Document 
Analysis, {\nineit Proceedings of SIGIR}, 1994
\bibitem{karlgren} J.~Karlgren, D.~Cutting: Recognizing Text Genres with Simple 
Metrics Using Discriminant Analysis, {\nineit Proceedings of COLING}, Kyoto, 
1994
\bibitem{schueHB} U.~Kressel, J.~Sch\"urmann: Pattern Classification Techniques 
Based on Function Approximation, to appear in: P.S.P.~Wang, H.~Bunke (eds.), 
Handbook of Optical Character Recognition and Document Image Analysis, 1996
\bibitem{lewis} D.~Lewis: Feature Selection and Feature Extraction for Text 
Categorization, {\nineit Proceedings of Speech and Natural Language Workshop}, 
1992
\bibitem{riloff} E.~Riloff: Automatically Constructing a Dictionary for 
Information Extraction Tasks, {\nineit Proceedings of 11th National Conference 
on Artificial Intelligence}, 1993
\bibitem{yang92} Yiming Y.~Yang, Christopher G.~Chute: An Application of Least 
Squares Fit Mapping To Text Information Retrieval {\nineit Proceedings of 
SIGIR}, 1983
\bibitem{yang95} Yiming Y.~Yang: Noise reduction in a statistical approach to 
text categorization.  {\nineit Proceedings of SIGIR}, 1995

\end{thebibliography}
\end{document}